\documentclass[twoside,11pt]{article}

\usepackage{blindtext}

\usepackage[preprint]{jmlr2e}

\usepackage[utf8]{inputenc}           
\usepackage{amsmath,amsfonts}         
\usepackage{algorithm}
\usepackage{algpseudocode}            
\usepackage[section]{placeins} 
\usepackage{multirow}

\usepackage{subcaption}               
\usepackage{listings}                 
\usepackage{xcolor}                  
\usepackage{booktabs}                 
\usepackage[capitalize,noabbrev]{cleveref} 
\usepackage{bm} 
\newcommand{\mat}[1]{\mathbf{#1}}     
        
\usepackage{xcolor}

\color{black}

\usepackage{orcidlink}
\usepackage[normalem]{ulem} 

\newcommand\rsout{\bgroup\markoverwith{\textcolor{red}{\rule[0.5ex]{2pt}{0.4pt}}}\ULon}

\usepackage{lastpage}

\ShortHeadings{Greedy-Gnorm for Attention Head Pruning}{Guo and Sheridan}
\firstpageno{1}

\begin{document}

\title{Greedy-Gnorm: A Gradient Matrix Norm-Based Alternative to Attention Entropy for Head Pruning}

\author{\name Yuxi Guo\,\orcidlink{0009-0005-1053-3577}
        \email yuxiguo03@gmail.com \\
       \addr SWUFE-UD Institute of Data Science\\
       Southwestern University of Finance and Economics\\
       Chengdu, Sichuan, China
       \AND
       \name Paul Sheridan\,\orcidlink{0000-0002-5484-1951} \email paul.sheridan.stats@gmail.com \\
       \addr School of Mathematical and Computational Sciences\\
       University of Prince Edward Island\\
       Charlottetown, PE, Canada}

\editor{TBD}

\maketitle

\begin{abstract}
Attention head pruning has emerged as an effective technique for transformer model compression, an increasingly important goal in the era of Green AI. However, existing pruning methods often rely on static importance scores, which fail to capture the evolving role of attention heads during iterative removal. We propose Greedy-Gradient norm (Greedy-Gnorm), a novel head pruning algorithm that dynamically recalculates head importance after each pruning step. Specifically, each head is scored by the elementwise product of the $\ell_2$-norms of its Q/K/V gradient blocks, as estimated from a hold-out validation set and updated at every greedy iteration. This dynamic approach to scoring mitigates against stale rankings and better reflects gradient-informed importance as pruning progresses. Extensive experiments on \textsc{BERT}, \textsc{ALBERT}, \textsc{RoBERTa}, and \textsc{XLM-RoBERTa} demonstrate that Greedy-Gnorm consistently preserves accuracy under substantial head removal, outperforming attention entropy. By effectively reducing model size while maintaining task performance, Greedy-Gnorm offers a promising step toward more energy-efficient transformer model deployment.
\end{abstract}


\section{Introduction}
Transformer architectures have become foundational in natural language processing, serving as the backbone of contemporary large language models. While these models achieve remarkable accuracy across a wide range of tasks, their computational demands and parameter redundancy raise concerns about energy efficiency, deployment cost, and feasibility on resource-constrained devices~\citep{Strubell2019}. The attention mechanism, in particular, is computationally expensive and potentially overparameterized, motivating research into techniques for reducing redundancy with minimal loss of performance.

A prominent strategy is attention head pruning, which reduces model size by removing those heads that are determined to be of least importance~\citep{Voita2019}. This approach complements other such compression methods as model quantization~\citep{Wang2024} and knowledge distillation~\citep{Muralidharan2024, Sreenivas2024}, and can be integrated with them to achieve additional efficiency gains. By directly targeting redundant heads, pruning can reduce both memory footprint and computation time, thereby enabling more sustainable and deployable transformer models~\citep{Wang2021a, Lagunas2021, Shim2021}.

Existing pruning methods have key limitations. A commonly used approach is attention entropy (AE), which assumes that heads with lower entropy produce more concentrated, or peaked, attention distributions (i.e., greater probability mass is placed on a few key positions) and are therefore more important~\citep{Voita2019, Wang2020}. However, AE is problematic when attention becomes diffused over long input sequences, as the resulting small probability values make the entropy computation vulnerable to numerical underflow. Moreover, most existing methods rely on static importance scores that are computed once before pruning~\citep{Michel2019, Hao2020, McCarley2021}. Since gradients and model dynamics evolve after each head removal, such static scores may become stale, potentially leading to suboptimal pruning decisions.

In this paper, we introduce Greedy-Gnorm, a dynamic, gradient-driven pruning strategy that recalculates transformer model head importance after every pruning step. Each head is scored by Gnorm, defined as the elementwise product of the $\ell_2$-norms of the model's Q/K/V gradient matrices, estimated on a hold-out validation set. By recomputing scores at each greedy iteration, Greedy-Gnorm adapts to evolving gradients and avoids the pitfalls of static ranking. When computing AE values, we impose a small lower bound $\varepsilon>0$ on each attention probability to prevent $\log(0)$ numerical underflow errors while leaving the pruning order essentially in tact. We refer to this technique as the $\varepsilon$-rectified entropy variant. To avoid numerical instability in AE-based methods, we compute entropy using an $\varepsilon$-rectified distribution, where a small constant $\varepsilon$ is added to each attention term to prevent taking the logarithm of $0$. This stabilization does not alter the head-importance ranking, so the resulting pruning order remains identical to the non-rectified case.

We validate Greedy-Gnorm on four widely used transformer model families: \textsc{BERT}~\citep{Devlin2018}, \textsc{ALBERT}~\citep{Lan2019}, \textsc{RoBERTa}~\citep{Liu2019}, and \textsc{XLM-RoBERTa}~\citep{Conneau2019}. Our experiments show that Greedy-Gnorm consistently preserves task accuracy under substantial head removal and exhibits stable behavior across architectures. Compared with AE and a random pruning baseline, our method yields more reliable pruning trajectories. After pruning, \textsc{BERT}, \textsc{ALBERT}, \textsc{RoBERTa}, and \textsc{XLM-RoBERTa} retain high accuracy, showing only modest declines relative to their unpruned counterparts. For \textsc{BERT}, we retain approximately $20\%$ of the heads while maintaining $90.08\%$ accuracy, compared to $96.82\%$ accuracy before pruning. Compared with the AE scoring approach and random baseline, Greedy-Gnorm produces smoother, more reliable pruning trajectories and achieves higher accuracy at equivalent pruning rates. 

The remainder of this paper is organized as follows. \cref{sec:background} introduces transformer model notation, surveys pruning and efficiency methods for these models, and situates our work within this context. \cref{sec:methodology} formalizes the Greedy-Gnorm algorithm, with emphasis on the method’s prune–recompute procedure and the $\varepsilon$-rectified entropy variant used to ensure numerical stability during head ranking. \cref{sec:experiments} details datasets, models, baselines, and reports results and ablations. Finally, \cref{sec:limitations} discusses limitations and future directions, and \cref{sec:conclusion} provides an overview of our contributions.

\section{Background}
\label{sec:background}
This section establishes the notation for transformer models adopted in the remainder of the paper and summarizes essential background on attention head pruning.

\subsection{Transformer Model Notation}
\label{sc:transformer-notation}
Let $\mathcal{X}$ denote a set of input sentences.  
Each sentence $x \in \mathcal{X}$ has a length of $t_x$ tokens after subword tokenization. Let the input be a tokenized sequence $x=(x_1,\ldots,x_{t_x})$ after subword tokenization for a sentence. For an input sentence $x$, each token $x_i$ is an index in the vocabulary $\mathcal{V}$ with $|\mathcal{V}|=V$ (i.e., $x_i\in\{1,\ldots,V\}$). 

Consider a transformer model $\mathcal{M}$ with $L>0$ layers, each containing $H>0$ heads, yielding a total of $N = L \times H$ heads. We assume the model has been trained on a set $\mathcal{X}$. Let $d_\mathcal{M}>0$ denote the dimension of the vector (i.e., embedding size). We assume that input and output sizes are held constant at $d_\mathcal{M}$ across the $L$ layers. An embedding layer is a matrix $E\in\mathbb{R}^{V\times d_{\mathcal M}}$ that maps each token to a vector $E[x_i]\in\mathbb{R}^{d_{\mathcal M}}$. Stacking these row vectors yields the embedding sequence in a matrix $X\in\mathbb{R}^{t_x\times d_{\mathcal M}}$ (i.e., a matrix with one $d_{\mathcal M}$-dimensional row per token). We assume the model preserves this width across all $L$ layers, so the input and output of each layer are in $\mathbb{R}^{t_x\times d_{\mathcal M}}$. 

In multi-head attention, with $H$ attention heads per layer, each global projection (i.e., the weights matrices $W_Q$, $W_K$, $W_V$) in each layer is split into $H$ head-specific projection blocks. We conceptually partition the projection output channels into $H$ equal parts. Let~$d_h$ denote the per-head output width. That is, the number of columns in a single head’s $Q/K/V$ projection block. Thus each head uses a matrix of dimension $d_{\mathcal M}\!\times\! d_h$, where $d_h$ is a head-wise constant given by $d_h = d_{\mathcal M}/H$. This choice ensures that concatenating the $H$ heads produces an output of width $H d_h = d_{\mathcal M}$ (i.e., the multi-head attention block preserves the model width, so the output has the same dimensionality as the input). The full projection can thus be viewed as the concatenation of $H$ head-specific blocks.

The projection matrices for head $h$ $(1\leq h \leq H)$ in layer $\ell$ $(1\leq\ell\leq L)$ are given by
\begin{equation}
Q^{(\ell,h)} \;=\; X\,W_Q^{(\ell,h)},\quad
K^{(\ell,h)} \;=\; X\,W_K^{(\ell,h)},\quad
V^{(\ell,h)} \;=\; X\,W_V^{(\ell,h)},
\end{equation}
with $W_Q^{(\ell,h)},W_K^{(\ell,h)},W_V^{(\ell,h)}\in\mathbb{R}^{d_{\mathcal{M}}\times d_h}$, and hence $Q^{(\ell,h)},K^{(\ell,h)},V^{(\ell,h)}\in\mathbb{R}^{t_x\times d_h}$.
The row-wise attention matrix is defined as
\begin{equation} \label{eq:attention-matrix}
A^{(\ell,h)}(x)\;=\;\mathrm{softmax}\!\left(\frac{Q^{(\ell,h)}K^{(\ell,h)\top}}{\sqrt{d_h}}\right)\in[0,1]^{t_x\times t_x},
\end{equation}
where $\sum_{j=1}^{t_x} a^{(\ell,h)}_{ij}(x)=1$ for all $i$. The head output is $Z^{(\ell,h)}=A^{(\ell,h)}V^{(\ell,h)}\in\mathbb{R}^{t_x\times d_h}$, and the multi-head output of layer $\ell$ is $\mathrm{Concat}_h\, (Z^{(\ell,h)})\,W_O^{(\ell)}$ with $W_O^{(\ell)}\in\mathbb{R}^{(H d_h)\times d_\mathcal{M}}$, where $\mathrm{Concat}_h(\cdot)$ denotes the concatenation of all head outputs 
along the feature dimension. We will refer to the head-specific parameter blocks $W_Q^{(\ell,h)},W_K^{(\ell,h)},W_V^{(\ell,h)}$ when defining gradient-based scores. The gradients of these matrices have the same dimensions as the corresponding blocks.

\subsection{Related Work}
\label{sec:related-work}
Here we review head-importance criteria and pruning strategies relevant to our approach. We begin with preliminaries and AE scoring and its numerical caveats, then cover gradient-based measures and the need for step-wise recomputation, and finally summarize depth-wise, width-wise, and length-wise pruning families and tooling.

\subsubsection{Attention Entropy}
For a sentence $x \in \mathcal{X}$ with length $t_x$, consider the attention-score matrix $A^{(\ell,h)}(x) \in [0,1]^{t_x \times t_x}$ of head $(\ell,h)$ as defined in Eq.~\eqref{eq:attention-matrix}. Writing its entries as $a_{ij}^{(\ell,h)}(x)$, the expected AE is defined as
\begin{equation}
\mathbb{E}\big[AE(\ell,h)\big]
= -\frac{1}{|\mathcal{X}|}\sum_{x\in\mathcal{X}}\frac{1}{t_x}\sum_{i=1}^{t_x}\sum_{j=1}^{t_x}
a_{ij}^{(\ell,h)}(x),\log a_{ij}^{(\ell,h)}(x).
\label{eq:ae}
\end{equation}
AE serves as a proxy for the “focus” of an attention head, quantifying how concentrated or diffuse its attention distribution is. However, AE disregards gradients and can be numerically brittle for long sequences, where many small $a_{ij}^{(\ell,h)}$ values make $-a\log a$ unstable under finite precision, occasionally leading to underflow or NaNs.

\subsubsection{Gradient-based Importance}
Gradient-aware scores use the loss derivatives $\partial \mathcal{L}(x)/\partial A^{(\ell,h)}(x)$ with respect to a head’s attention. Let $A^{(\ell,h)}(x)\in[0,1]^{t_x\times t_x}$ be the attention of head $(\ell,h)$, and collect all heads in layer $\ell$ as
\begin{equation}
A^{(\ell)}(x) = \big[A^{(\ell,1)}(x),\ldots,A^{(\ell,H)}(x)\big].
\end{equation}
The attribution score for head $(\ell,h)$ is computed using a gradient-based self-attention analysis method~\citep{Hao2020}. Self-attention attribution for head $(\ell,h)$ is
\begin{align}
\mathrm{Attr}^{(\ell,h)}(x)
&= A^{(\ell,h)}(x)\ \odot\ \int_{0}^{1}
\nabla_{A^{(\ell,h)}} F\!\big(x;\,\alpha\,A^{(\ell)}(x)\big)\,d\alpha \\
&\approx \frac{1}{m}\,A^{(\ell,h)}(x)\ \odot\ \sum_{k=1}^{m}
\nabla_{A^{(\ell,h)}} F\!\big(x;\,\tfrac{k}{m}\,A^{(\ell)}(x)\big)\in\mathbb{R}^{t_x\times t_x},
\end{align}
where $F(\cdot;\cdot)$ is a differentiable scalarization of the model output and $\odot$ denotes the Hadamard product. Then define head importance by
\begin{equation}
I^{(\ell,h)} = \mathbb{E}_{x\sim X}\!\big[\max(\mathrm{Attr}^{(\ell,h)}(x))\big].
\end{equation}
where $F(\cdot)$ is a differentiable scalarization of the model output and $\odot$ is the Hadamard product. Taylor-based importance (TIS)~\citep{Zhong2024} can be written using the inner product
\begin{equation}
\mathrm{TIS}^{(\ell,h)} \;=\; \mathbb{E}_{x}\!\left|A^{(\ell,h)}(x)\ \frac{\partial \mathcal{L}(x)}{\partial A^{(\ell,h)}(x)}\right|.
\end{equation}
Beyond these, differentiable subset selection~\citep{Li2023} and Shapley-based head valuation~\citep{Held2022} have been explored. However, once any head is pruned, gradients and representations shift, so importance must be re-evaluated at each step; otherwise scores become stale and may mislead subsequent choices~\citep{Michel2019, McCarley2021, Parnami2021, Shim2021}. This motivates our greedy, recompute-as-you-pruning procedure.

\subsubsection{Other Transformer Pruning Strategies}
Transformer compression has been explored along three structural dimensions: depth, width, and sequence length, each targeting different computational bottlenecks.

One line of work focuses on reducing model depth. For example, static layer removal and sensitivity-driven pruning reduces depth with post-hoc fine-tuning~\citep{McCarley2021}. Dynamic policies skip layers per input to trade quality for cost~\citep{Bapna2020}. Recent Mixture-of-Depths dynamically allocates compute across layers~\citep{Raposo2024}, and layer importance can be optimized via NAS/analysis~\citep{Klein2024, Zhang2024}. Related architectural simplifications include average-attention networks~\citep{Zhang2018}, reordering sublayers~\citep{Press2020}, and light-weight stacks~\citep{Mehta2021}.

Another research direction explores pruning along the width dimension, primarily within layers. Head pruning ranges from early evidence of redundancy~\citep{Michel2019, Voita2019} to structured gating~\citep{Shim2021}, search-based A*~\citep{Parnami2021}, differentiable subsets~\citep{Li2023}, Shapley valuations~\citep{Held2022}, and sparse attention with cascaded token-then-head sparsity~\citep{Wang2021a}. Feedforward neural network/channel/filter pruning targets the dominant MLP cost, often with block structure for hardware efficiency~\citep{Lagunas2021, Liu2021, Yu2022}. Tooling such as \textsc{TextPruner} supports practical pipelines~\citep{Yang2022}.

A complementary body of work targets sequence length reduction. Token-level methods prune, merge, skip, or drop tokens to shorten effective context; policies can be static or dynamic/adaptive~\citep{Lee2022,Lee2025}. System-oriented work reduces KV-cache traffic via selective fetching or cache designs~\citep{He2024,Sun2024}, complementary to head pruning.

\section{Methodology} \label{sec:methodology}
In the section, we present the Greedy-Gnorm algorithm and introduce a simple technique designed to prevent underflow errors in AE calculations.

\subsection{The Greedy-Gnorm Algorithm}
Our goal is to prune a transformer model of redundant attention heads while preserving task accuracy. To this end, we propose Greedy-Gnorm, a greedy pruning scheme that leverages importance scores reflecting the current state of a partially pruned model.  Each iteration consists of two steps: (1) compute the current Greedy-Gnorm score matrix based on a current gradient matrix using backpropagation, and (2) prune the attention head with the lowest score. This prune-recompute cycle continues until either a pruning budget is exhausted or an accuracy-based stopping criterion is satisfied, thereby avoiding the staleness of one-shot scores and producing more stable pruning trajectories. In what follows, we formalize notation, describe the construction of the scoring matrix, and analyze the algorithmic complexity of the proposed method.

\subsubsection{Greedy-Gnorm Algorithm High-level Description}
The Greedy-Gnorm routine is described in \cref{alg:greedy-gnorm}. We adopt the notation of Section~\ref{sc:transformer-notation}. Consider a transformer $\mathcal{M}$, given input $X$, the output of original $\mathcal{M}$ is $F_0(X)$. Let $M\in\{0,1\}^{L\times H}$ be the mask (1=kept, 0=pruned).

\begin{algorithm}[!htbp]
\caption{The Greedy-Gnorm algorithm.}
\label{algo1}\label{alg:greedy-gnorm}
\begin{algorithmic}[1]
\Require Initial model $F_0$; the set $\mathcal{X}$; number of layers $L$; heads per layer $H$; $M \in \{0,1\}^{L\times H}$ and initialize $M \gets\mathbf{1}_{L\times H}$; number of total heads $N \gets L\times H$.
\Ensure Pruned model $F$ and accuracies of each pruned model.

\State $F \gets F_0$ \Comment{current model}
\State $M \gets \mathbf{1}_{L\times H}$ \Comment{mask (1 = keep, 0 = pruned)}
\For{$n=0$ to $N-1$} \Comment{greedy re-computation loop}
  \State $(G_{Q(n)}, G_{K(n)}, G_{V(n)}) \gets \Call{ComputeGnorm}{F, \mathcal{X}, M}$
  \State $S(n) \gets G_{Q(n)} \odot G_{K(n)} \odot G_{V(n)}$ \Comment{elementwise product, $S\in\mathbb{R}^{L\times H}$}
  \State $(\ell^\star, h^\star) \gets \arg\min\{\, S(n)_{\ell h}\mid M_{\ell h}=1 \,\}$ \Comment{least-important head}
  \State $F \gets \Call{PruneHead}{F, \ell^\star, h^\star}$ \Comment{apply head-gating or equivalent}
  \State $M_{\ell^\star h^\star} \gets 0$ \Comment{update mask}
  \State \Call{GetAccuracy}{$F, \mathcal{X}$} \Comment{get the accuracy of model on $\mathcal{X}$}
\EndFor
\State \Return $(F, M)$
\end{algorithmic}
\end{algorithm}
Following \cref{alg:greedy-gnorm}, we describe the full pruning procedure step by step and explain how each quantity is computed. 
At initialization (lines~1–2), the algorithm loads the pretrained transformer $F_0$ and sets up the binary mask $M \in \{0,1\}^{L\times H}$, 
where $M_{\ell h}=1$ indicates that the head $(\ell,h)$ is active. 
The model $F$ is iteratively updated under this mask throughout the pruning process.

The outer loop (line~3) performs $N=L\times H$ pruning iterations, each removing one attention head. 
At the beginning of each iteration, the subroutine \textsc{ComputeGnorm} (line~4) computes three gradient-norm matrices 
$G_{Q(n)}$, $G_{K(n)}$, and $G_{V(n)}$, corresponding respectively to the query, key, and value projection blocks 
$W_Q^{(\ell,h)}$, $W_K^{(\ell,h)}$, and $W_V^{(\ell,h)}$. 

Line~5 combines these matrices through an elementwise product, which captures the joint strength of the three attention projections. 
$S(n)$ therefore acts as a unified importance score matrix: heads with smaller values of $S(n)_{\ell h}$ 
are considered less influential.

Next, line~6 identifies the least-important active head $(\ell^\star, h^\star)$ by finding the minimum entry in $S(n)$ over unpruned positions ($M_{\ell h}=1$). The selected head is removed from the model in line~7 via the \textsc{PruneHead} operation, which structurally removes the corresponding attention head using the \textsc{TextPruner} framework (see \Cref{app:collapsed-gradient} for implementation details and discussion of gradient collapse). 
The mask $M$ is updated accordingly in line~8 to record this removal.

After pruning, the model’s performance is immediately re-evaluated on $\mathcal{X}$ (line~9) using \textsc{GetAccuracy}. This greedy prune–recompute cycle continues until all heads have been ranked and sequentially pruned. The final output $(F, M)$ (line~10) contains the fully pruned model and the final mask, where $M=\mathbf{0}_{L\times H}$ after all heads have been removed, indicating that the pruning process has ranked and eliminated every attention head.

\subsubsection{Technical Details}
To execute \cref{alg:greedy-gnorm}, we need the gradient-norm matrices $G_{Q(n)}$, $G_{K(n)}$, and $G_{V(n)}$, each summarizing the average magnitude of parameter gradients for query, key, and value projections across all layers and heads. The elementwise product $S(n) = G_{Q(n)} \odot G_{K(n)} \odot G_{V(n)}$ thus aggregates their joint contribution and serves as a unified importance score matrix. In the following section, we formally define $G_{Q(n)}$, $G_{K(n)}$, $G_{V(n)}$, and $S(n)$. 

We next define the Gnorm score, which quantifies the importance of each attention head based on its gradient matrix norms. For layer $\ell$ and head $h$, $W_Q^{(\ell,h)},W_K^{(\ell,h)},W_V^{(\ell,h)}$ are the head-specific projection blocks with dimensions as defined in~\Cref{sc:transformer-notation}. Write $q^{(\ell,h)}$ for the collection of scalars in $W_Q^{(\ell,h)}$, with $q^{(\ell,h)}_{a,b}$ its entry at $a$-th row and $b$-th column in the $W_Q^{(\ell,h)}$; analogously $k^{(\ell,h)}$ and $v^{(\ell,h)}$ for $W_K^{(\ell,h)}$ and $W_V^{(\ell,h)}$. 

After pruning $n$ heads, the current model is $F_n(\cdot)$ and we define our gradient matrix for Q of $\ell$'th layer, $h$'th head and denote the Euclidean norm matrix for each head as follows
\begin{eqnarray}
\mat G^{(\ell,h)}_{q(n)}(X)
&=& \nabla_{\,W_Q^{(\ell,h)}} \,\big\| F_n(X) \big\|
\;=\; \left[\,\frac{\partial \big\| F_n(X) \big\|}{\partial q^{(\ell,h)}_{a,b}}\,\right]_{1 \le a \le d_{\mathcal M},\, 1 \le b \le d_h} \\
\big\| \mat G^{(\ell,h)}_{q(n)}(X) \big\|
&=& \sqrt{\displaystyle \sum_{a=1}^{d_{\mathcal M}}\sum_{b=1}^{d_h}\left(\frac{\partial \big\| F_n(X) \big\|}{\partial q^{(\ell,h)}_{a,b}}\right)^{\!2}} \\
\mat G_{q(n)}
&=& \left(\, \big\| \mat G^{(\ell,h)}_{q(n)}(X) \big\| \,\right)_{1 \le \ell \le L,\; 1 \le h \le H}
\in \mathbb{R}^{L \times H} \\
&=& \left(\, \sqrt{\displaystyle \sum_{a=1}^{d_{\mathcal M}}\sum_{b=1}^{d_h}\left(\frac{\partial \big\| F_n(X) \big\|}{\partial q^{(\ell,h)}_{a,b}}\right)^{\!2}} 
\,\right)_{1 \le \ell \le L,\; 1 \le h \le H}
\in \mathbb{R}^{L \times H}. \label{eq:gnorm-entry-brace} 
\end{eqnarray}
where $\mat G^{(\ell,h)}_{q(n)}$ means the gradient matrix of Q weights at $\ell^{th}$ layer $h^{th}$ head. Correspondingly, $\mat G^{(\ell,h)}_{k(n)}$ and $\mat G^{(\ell,h)}_{v(n)}$ mean the gradient matrix of K and V weights at in $\ell^{th}$ layer $h^{th}$ head. $\mat G_q(n)$ ,$\mat G_k(n)$ and $\mat G_v(n)$ are Euclidean norm matrices of Q, K and V in each head. We use $\|F_n(X)\|$ as a scalarization of the model output (e.g., logits $\ell_2$ norm). Note that any differentiable scalar objective (e.g., task loss) can be used.

Based on different inputs in datasets, we need to consider the expectations of the gradient matrix norms:
\begin{eqnarray}
\mat G_{Q(n)}
&=& \mathbb{E}_{X}\big[\mat G_{q(n)}\big] \\
&=& \left(\, \mathbb{E}_{X}\big[\big\|\mat G^{(\ell,h)}_{q(n)}(X)\big\|\big] \,\right)_{1 \le \ell \le L,\; 1 \le h \le H}
\in \mathbb{R}^{L \times H} \\ 
&=& \left(\, \frac{1}{|\mathcal X|}\sum_{X}
\sqrt{\displaystyle\sum_{a=1}^{d_{\mathcal M}}\sum_{b=1}^{d_h}
\left(\frac{\partial \big\|F_n(X)\big\|}{\partial q^{(\ell,h)}_{a,b}}\right)^{\!2}} \,\right)_{1 \le \ell \le L,\; 1 \le h \le H}
\in \mathbb{R}^{L \times H}. \label{eq:gnorm-expect-entry}
\end{eqnarray}
$\mat G_{Q(n)}$ is the matrix of expected norms for $L\times H$ heads, and analogously for $\mat G_{K(n)}$ and $\mat G_{V(n)}$. We take the expectation over a set $\mathcal{X}$ of sentences, where each $X$ denotes the embedding result of a single sentence $x\in\mathcal{X}$ input to the transformer model. 

Here, $G_{Q(n)}(\ell,h)$ denotes the entry in the $\ell$-th row and $h$-th column of $\mat G_{Q(n)}$, with $G_{K(n)}(\ell,h)$ and $G_{V(n)}(\ell,h)$ defined analogously:
\begin{equation}
G_{Q(n)}(\ell,h) = \mathbb{E}_{X}\!\left[\big\|\mat G^{(\ell,h)}_{q(n)}(X)\big\|\right],
\quad
\end{equation}
After pruning $n$ heads, we define the Gnorm score matrix $\mat S(n)\in\mathbb{R}^{L\times H}$ as the elementwise product:
\begin{equation}
\mat S(n) \;=\; \mat G_{Q(n)} \odot \mat G_{K(n)} \odot \mat G_{V(n)},\,\mat S(n) \in \mathbb{R}^{L \times H}.
\label{eq:gnorm}
\end{equation}
At each greedy step, we prune the head with the smallest $S(n)_{ij}$ and then recompute gradients and $\mat S(n{+}1)$.
So $\mat S(n)$ tells that after pruning $n$ heads, which head is the least important to do next pruning. 

At each greedy step, we prune the head with the smallest $S(n)_{ij}$ and then recompute gradients and $\mat S(n{+}1)$. 
Thus, $\mat S(n)$ indicates, after pruning $n$ heads, which of the remaining heads is least important for the next step.

The algorithm proceeds iteratively, ranking all heads by their Gnorm scores until every head has been evaluated or pruned. 
In practice, pruning can be stopped earlier when accuracy begins to drop sharply, which marks the empirical “inflection point” beyond which further pruning causes disproportionate degradation. 
This point provides a principled trade-off between compactness and performance.

\subsection{Time Complexity}
Let the calibration set $\mathcal{X}$ be divided into $B$ mini-batches for gradient computation. In each greedy iteration, \textsc{ComputeGnorm} performs a full backward pass for every batch in each layer to obtain the $Q/K/V$ gradients of all heads, yielding a per-iteration cost of $\textsc{ComputeGnorm} = \Theta(L\,B)$. Because pruning proceeds iteratively, recomputing gradients after each head removal, the total number of iterations is proportional to $N$. Hence, the overall computational complexity of Greedy-Gnorm is $\mathcal{O}(N\,L\,B)$. The scoring and selection steps, which form the score tensor $S$ and locating $\arg\min S$, require only $\mathcal{O}(N)$ time per iteration and are negligible compared to the cost of backpropagation.

\subsection{A Technique for Underflow Error Avoidance}
AE scores heads via $-\sum_i a_i\log a_i$ averaged over tokens and examples. On long sequences, attention may become diffuse resulting in many entries $a_i$ approaching machine zero. This leads to $\log a_i$ underflow errors (or $a_i=0$ after softmax rounding), producing $\log(0)$ and NaNs that corrupt averages and break pruning order. Although Greedy-Gnorm itself is gradient-based, we employ AE as a baseline. Hence we need a numerically stable way of computing AE scores.

Let $\mathbf{a}=(a_1,\dots,a_n)$ denote a row of a head’s attention matrix (a probability vector with $\sum_i a_i=1$, $a_i\ge 0$).
Information entropy is
\begin{equation}
A(\mathbf{a}) = -\sum_{i=1}^n a_i \log a_i .
\label{eq:entropy-A}
\end{equation}
To avoid underflow we consider two $\varepsilon$-rectified variants with a small $\varepsilon>0$ (chosen so that $a_i+\varepsilon<1$):
\begin{align}
B(\mathbf{a}) &= -\sum_{i=1}^n a_i \log\!\big(a_i+\varepsilon\big), \label{eq:entropy-B} \\
C(\mathbf{a}) &= -\sum_{i=1}^n \big(a_i+\varepsilon\big)\log\!\big(a_i+\varepsilon\big). \label{eq:entropy-C}
\end{align}
The function $B$ clips the $log$ argument away from $0$ while keeping the original weights $a_i$, removing $\log(0)$ but leaving the averaging weights unchanged. The function $C$ also shifts the weights (from $a_i$ to $a_i+\varepsilon$), making the entropy itself less sensitive to tiny entries and thus more numerically stable. For sufficiently small $\varepsilon$, both act as smooth perturbations cof $A$. In practice we adopt $C$ for stronger stability (no NaNs) while preserving head rankings for typical $\varepsilon$ (see \cref{app:epsilon-discussion} for discussion). 

We choose $C$ for rectification for two reasons: (i) it fully removes $\log(0)$ and (ii) it remains order-preserving for small $\varepsilon$. Let $F(\mathbf{a})=\sum_i a_i\log a_i$. For two distributions $\mathbf{x},\mathbf{y}$ with $F(\mathbf{y})-F(\mathbf{x})=\delta>0$, consider $\mathbf{x}'=\mathbf{x}+\varepsilon\mathbf{1}$, $\mathbf{y}'=\mathbf{y}+\varepsilon\mathbf{1}$. Along the line segment joining $\mathbf{x}$ and $\mathbf{y}$, the directional derivative involves $\nabla F(\mathbf{a})=(\log a_i+1)_i$. Continuity implies that for sufficiently small $\varepsilon$, the sign of the line integral remains unchanged and thus the ordering remains unchanged:
\begin{equation}
\delta'=F(\mathbf{y}')-F(\mathbf{x}')=\int_{\mathbf{x}'}^{\mathbf{y}'}\nabla F(\mathbf{a})\!\cdot\! d\mathbf{l} \;>\; 0.
\end{equation}
However, for $B(\mathbf{a})$, it is uncertain to find whether the pruning order changes. At least, the pruning order induced by AE is preserved under $C(\mathbf{a})$ while eliminating underflow/NaN in practice.

\section{Experiments}
\label{sec:experiments}
In this section, we present empirical evidence demonstrating that our proposed attention head pruning method outperforms both the AE approach and a random pruning baseline.

\subsection{Setup}
\label{sec:setup}
We design our experiments to evaluate whether Greedy-Gnorm effectively identifies and preserves the most important attention heads under varying pruning budgets, while maintaining competitive accuracy on selected classification tasks. Conceptually, our goal is to assess the degree to which dynamically updated gradient norms yield more stable pruning trajectories as compared with static and random baselines. In each experiment, we progressively removes heads according to a ranking criterion, and measure the resulting performance on downstream classification tasks.

\begin{figure}[!htbp]
\centering
\includegraphics[width=1.0\linewidth]{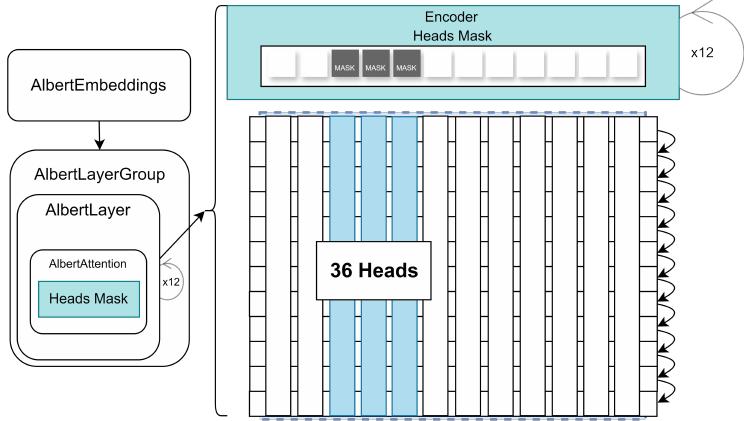}
\caption{\textsc{ALBERT} tied-head structure (appearing as a vertical band in the mask visualization). 
Head $h$ shares its parameters across all $L$ layers, gating a single mask entry disables the same head index in every layer. One mask position affects 12 heads across layers.}
\label{fig:albert}
\end{figure}

\paragraph{Models and tasks.}
We evaluate the Greedy-Gnorm algorithm on four pretrained transformer models, each paired with a distinct downstream task:  (1) \textsc{BERT} on financial sentiment classification, (2) \textsc{ALBERT} on Multi-Genre Natural Language Inference, (3) \textsc{RoBERTa} on tweet sentiment analysis, and (4) \textsc{XLM-RoBERTa} on language identification. In the case of \textsc{ALBERT}, parameter sharing ties all 12 attention layers, so one mask position simultaneously affects 12 heads, as illustrated in~\cref{fig:albert}. 

\paragraph{Pruning methods.}
We compare five pruning strategies to evaluate the role of dynamic gradient-based ranking:
\begin{description}
    \item[Greedy-Gnorm:] A method that dynamically updates gradient norms to estimate head importance during iterative pruning. Scores are recomputed after each pruning step from the current model state.
    \item[AE:] A static baseline that ranks heads by their AE. Scores are computed once on the unpruned model and then held fixed.
    \item[Inverse-AE:] A static baseline that prunes the most informative (i.e., high-entropy) heads first, providing a complementary perspective to AE.
    \item[Inverse-Gnorm:] A dynamic baseline that removes the most gradient-active heads first, thereby retaining the least important ones to illustrate the effectiveness of Greedy-Gnorm appraoch. If pruning in reverse Gnorm order (removing the highest-scoring heads first) causes a rapid accuracy collapse, this will validate that our score faithfully captures head importance.
    \item[Random pruning:] A baseline that discards heads uniformly at each pruning step, providing a capacity-matched control to isolate the effect of head selection from mere capacity reduction.
\end{description}
Together, these strategies allow us to test whether dynamically updated, gradient-informed scores yield more stable and effective pruning trajectories than static, heuristic, or random approaches.

\paragraph{Evaluation metrics.}
Model performance is evaluated using classification accuracy on the held-out test sets. For each pruning method, we plot the accuracy curve against pruning rate to visualize robustness. Additionally, we report model sizes (in megabytes) before and after pruning to quantify compression efficiency. All results are averaged over three independent runs to mitigate randomness.

\subsection{Results}
\label{sec:results}
We now present the empirical results and interpret the key findings from all experiments. Our analysis proceeds from verifying the gradient dynamics induced by pruning, to comparing Greedy-Gnorm against static and random baselines, and finally to examining the resulting model compression outcomes.

\begin{figure}[!htbp]
\centering
\includegraphics[width=1.0\linewidth]{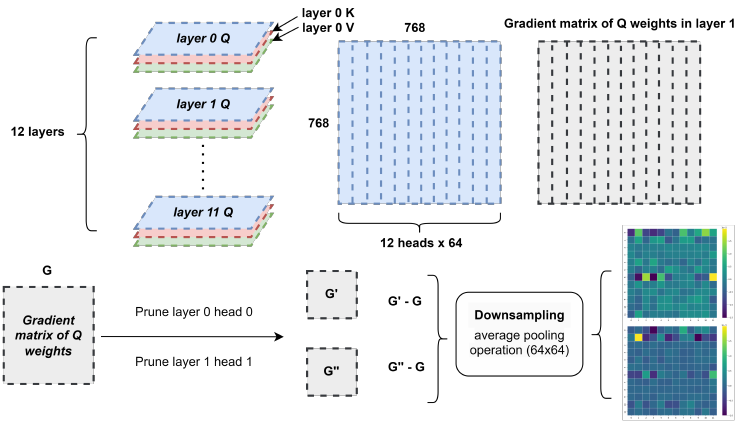}
\caption{Gradient changes after pruning (downsampled by $64\!\times\!64$ pooling from $768\!\times\!768$ to $12\!\times\!12$). Colors differ from $0$ across settings, indicating nonlocal gradient shifts.}
\label{fig:grad-heatmap}
\end{figure}

\subsubsection{Gradients Change After Pruning}
We begin by examining whether removing one attention head influences the gradients of others, a necessary condition for our dynamic scoring approach to be meaningful. As illustrated in~\cref{fig:grad-heatmap}, pruning a single head causes nontrivial changes in the Q-gradient matrices across both the pruned and unpruned layers. Even after $64\times64$ pooling (from $768\times768$ matrices), the heatmaps exhibit consistent deviations from $0$, indicating that pruning one head alters the gradient flow throughout the network. This observation empirically supports our assumption that head importance should be recomputed iteratively, since local pruning decisions induce global gradient redistribution.

\subsubsection{Main Findings}
\cref{tab:solutions} reports accuracy and model size before and after pruning across all architectures. It is clear that Greedy-Gnorm consistently reduces model size while preserving competitive accuracy. Greedy-Gnorm achieves substantial compression (e.g., $\approx 22.5\%$ size reduction on \textsc{BERT}) with competitive post-pruning accuracy. Similar trends hold for \textsc{RoBERTa} and \textsc{XLM-RoBERTa}, while \textsc{ALBERT}'s already compact design yields smaller absolute savings. 

\begin{table}[!htbp]
\centering
\caption{Greedy-Gnorm solutions. Percent reductions in both model size and accuracy are shown for readability.}
\label{tab:solutions}
\resizebox{\textwidth}{!}{
\begin{tabular}{lrrrrrr}
\toprule
 & \multicolumn{3}{c}{Accuracy (\%)} & \multicolumn{3}{c}{Size (MB)} \\
\cmidrule(lr){2-4} \cmidrule(lr){5-7}
Model & Before Pruning & After Pruning & Pct.\ Decrease & Before Pruning & After Pruning & Pct.\ Reduction \\
\midrule
BERT        & 96.82 & 90.08 & 6.97\% & 390.13  & 302.29  & 22.52\% \\
ALBERT      & 84.48 & 77.76 & 7.95\% & 44.58   & 42.33   & 5.05\%  \\
RoBERTa     & 87.80 & 86.40 & 1.59\% & 1355.60 & 1110.42 & 18.09\% \\
XLM-RoBERTa & 99.73 & 90.97 & 8.78\% & 1060.71 & 981.88  & 7.43\%  \\
\bottomrule
\end{tabular}
}
\end{table}

\begin{figure}[!htbp]
\centering
\includegraphics[width=1.0\linewidth]{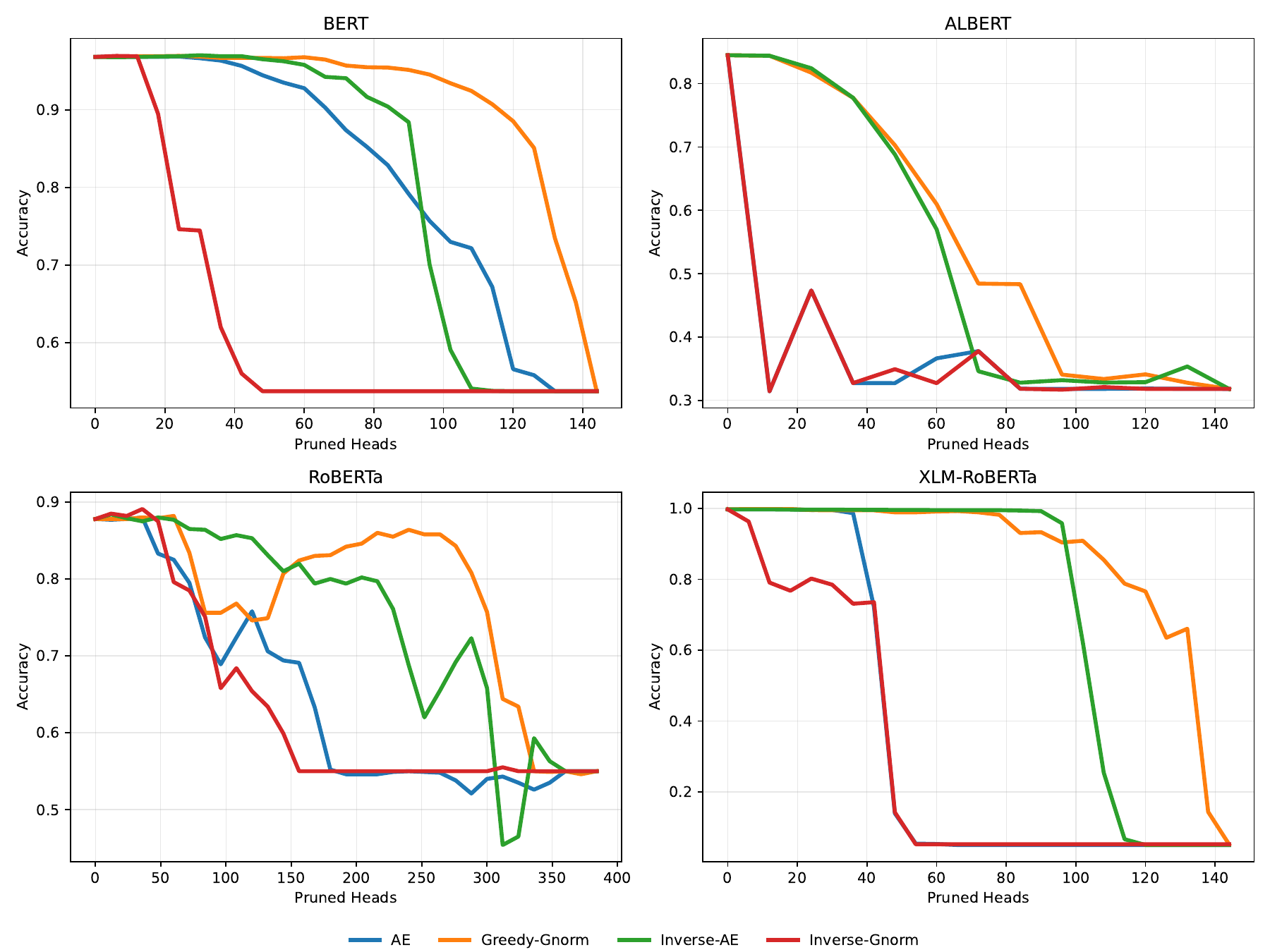}
\caption{Greedy-Gnorm vs. AE (and inverse variants). Greedy-Gnorm is more stable and preserves accuracy under deeper pruning.}
\label{fig:results}
\end{figure}

\paragraph{Greedy-Gnorm outperforms static baselines.}
We next compare Greedy-Gnorm with the dynamic (i.e., Inverse-Gnorm) and static (i.e., AE and Inverse-AE) baselines across all four models. As shown in \cref{fig:results}, Greedy-Gnorm exhibits a markedly smoother accuracy decay curves as pruning progresses. For \textsc{RoBERTa}, approximately 70\% of the heads suffice to retain near-original performance. A similar “useful head” fraction is observed for other architectures, suggesting that a relatively small subset of heads drives most of the task-relevant computation. In contrast, AE underperforms Greedy-Gnorm across all models. This is most notable on \textsc{BERT}, where Greedy-Gnorm preserves around 90\% accuracy with fewer than 20\% of heads retained.
These results demonstrate that dynamic gradient-based ranking provides more reliable pruning trajectories than static entropy-based criteria.

\paragraph{Validating importance scores via inverse pruning.}
To further validate the meaningfulness of the importance scores, we perform inverse pruning, in which the ranking order is deliberately reversed. This means that the most important heads, as estimated by each scoring method, are pruned first. This “reverse stress test” provides a diagnostic view of how well a given criterion distinguishes essential from redundant components. If the scoring function is informative, inverse pruning should lead to a rapid and monotonic performance collapse. Empirically, this is precisely what we observe. Inverse-Gnorm curves fall sharply after only modest pruning ratios, whereas Inverse-AE shows slower and more erratic degradation, reflecting noisier head importance estimates. The strong asymmetry between normal and Inverse-Gnorm trajectories confirms that our gradient-based metric captures functional salience rather than random variation.

\begin{figure}[!htbp]
\centering
\includegraphics[width=1.0\linewidth]{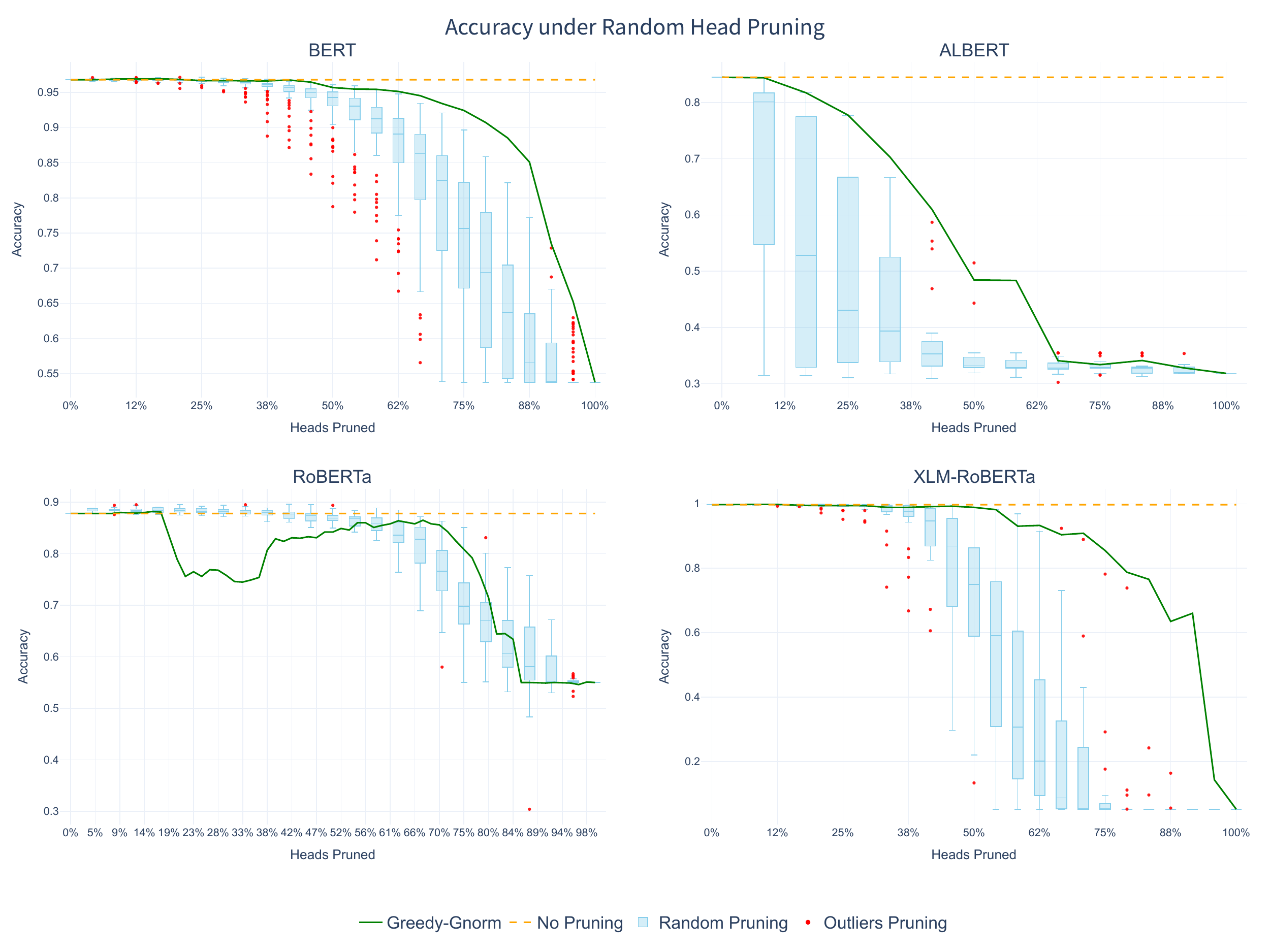}
\caption{Greedy-Gnorm versus random pruning across pruning rates. Boxplots summarize multiple random masks per rate, while the green curves show Greedy-Gnorm task accuracy. The dashed yellow line shows accuracy with no pruning.}
\label{figure5}
\end{figure}

\paragraph{Comparison with random pruning.}
We further benchmark Greedy-Gnorm against random head removal to quantify the contribution of structured selection. As illustrated in \cref{figure5}, across models, Greedy-Gnorm typically achieves the slowest overall accuracy decay. This slower decay indicates that the retained heads are genuinely informative and important. For \textsc{RoBERTa}, however, the abundance of redundant heads means that modest pruning can occasionally improve accuracy, yielding non-monotonic “rebound” segments. Nevertheless, the overall pruning outcomes on \textsc{RoBERTa} remain satisfactory. The discrepancy widens at higher pruning rates, reflecting that random removal often discards heads that remain functionally important. In contrast, Greedy-Gnorm’s score-driven masking avoids such destructive deletions, producing more stable and reproducible pruning behavior even when over 75\% of heads are pruned. This robustness underscores the value of gradient-informed importance estimation in guiding efficient compression.

\subsection{An Investigation of Pruning Solutions}

We next interpret the learned pruning patterns by visualizing the final head masks (light = pruned, dark = kept) across backbones and then drill down into BERT’s per-module parameter changes.

\begin{figure}[!htbp]
\centering
\includegraphics[width=1.0\linewidth]{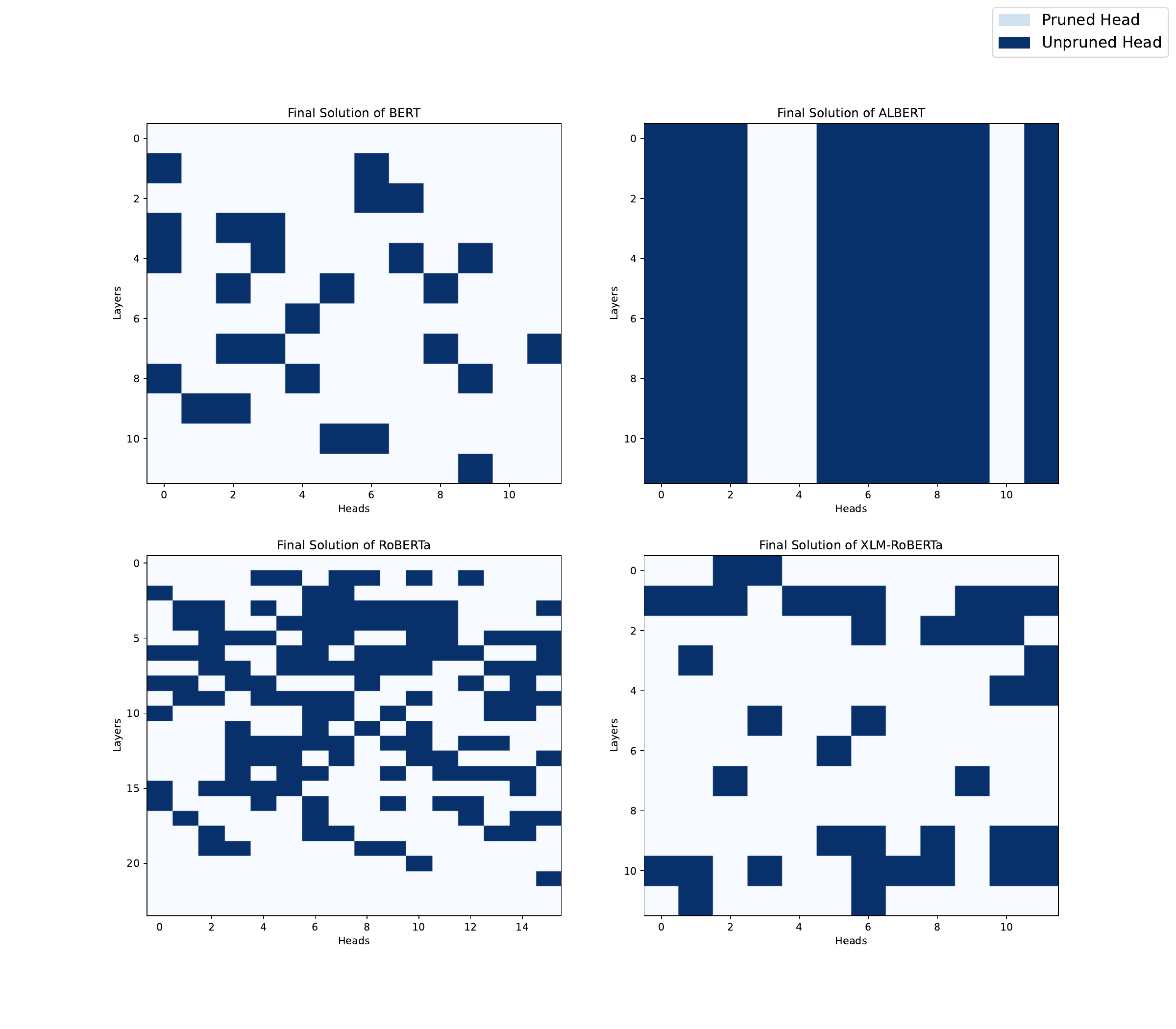}
\caption{Final pruning masks across models. BERT shows dispersed retention; ALBERT shows vertical bands due to parameter sharing; RoBERTa/XLM-RoBERTa display selective, layer-dependent retention.}
\label{fig:solutions}
\end{figure}

\cref{fig:solutions} visualizes the final head-retention masks for all backbones, with light cells indicating pruned heads and dark cells indicating retained heads. Each panel is arranged by layer (rows) and head index (columns). BERT exhibits dispersed retention across depth, ALBERT shows vertical bands due to parameter sharing (shared parameters induce identical keep/prune decisions across reused layers), while RoBERTa and XLM-RoBERTa display layer-dependent selectivity.

\cref{tab:bert-params} reports BERT’s per-module parameters before/after pruning. Most reduction concentrates in the Encoder block, as expected for head pruning (Embeddings, Pooler, and task head remain unchanged). Despite a sizable encoder shrinkage (i.e., from 85.05M to 62.03M parameters), post-pruning accuracy remains competitive (i.e., from 96.82\% to 90.08\%).

\begin{table}[!htbp]
\centering
\caption{BERT parameters before/after pruning with Greedy-Gnorm (MB=megabytes).}
\label{tab:bert-params}
\begin{tabular}{lrrrrrr}
\toprule
\multirow{2}{*} & \multicolumn{3}{c}{Before Pruning} & \multicolumn{3}{c}{After Pruning} \\
\cmidrule(lr){2-4} \cmidrule(lr){5-7}
{Layer} & Params & Share (\%) & MB & Params & Share (\%) & MB \\
\midrule
Model      & 102{,}269{,}955 & 100.00 & 390.13 & 79{,}244{,}355 & 100.00 & 302.29 \\
BERT & 102{,}267{,}648 & 100.00 & 390.12 & 79{,}242{,}048 & 100.00 & 302.28 \\
Embeddings & 16{,}622{,}592  & 16.25  & 63.41  & 16{,}622{,}592 & 20.98  & 63.41 \\
Encoder    & 85{,}054{,}464  & 83.17  & 324.46 & 62{,}028{,}864 & 78.28  & 236.62 \\
Pooler     & 590{,}592       & 0.58   & 2.25   & 590{,}592      & 0.75   & 2.25 \\
Classifier & 2{,}307         & 0.00   & 0.01   & 2{,}307        & 0.00   & 0.01 \\
Weight     & 2{,}304         & 0.00   & 0.01   & 2{,}304        & 0.00   & 0.01 \\
Bias       & 3               & 0.00   & 0.00   & 3              & 0.00   & 0.00 \\
\midrule
Accuracy   & \multicolumn{3}{c}{96.82\%} & \multicolumn{3}{c}{90.08\%} \\
\bottomrule
\end{tabular}
\end{table}

\section{Limitations and Future Work}
\label{sec:limitations}
Greedy-Gnorm trades stability for extra compute: each pruning step requires a backward pass on a small calibration set to refresh $Q/K/V$ gradients, making it costlier than one-shot scoring (roughly $\mathcal{O}(N\,B)$ for $N$ heads and $B$ batches). The approach also assumes the calibration distribution is representative of deployment; mismatch can mis-rank heads. Our implementation is head-only (no feedforward neural network or token pruning), so it cannot remove all redundancy. Gradients can be noisy for small batches or long contexts. 

Future work includes scaling to larger LLMs and additional architectures. Extending to multi-dimensional pruning that jointly selects heads and feedforward neural network width, and integrating structural removal so parameter cuts translate to latency gains. We also plan to combine Greedy-Gnorm with quantization and distillation for compounding compression, and to study alternative scalar objectives (e.g., task loss vs.\ logit norms).

\section{Conclusion}
\label{sec:conclusion}
We introduced Greedy-Gnorm, a head-pruning framework that scores heads using the elementwise product of the $\ell_2$ norms of their Q/K/V gradient matrices, and dynamically recomputes head importance after each removal. A simple mask-based expansion ensures that gradient matrix is comparable across pruning steps, and an $\varepsilon$-rectified approach to evaluating AE is employed as safeguard against numerical underflow. Numerical experiments with \textsc{BERT}, \textsc{ALBERT}, \textsc{RoBERTa}, and \textsc{XLM-RoBERTa} demonstrate that Greedy-Gnorm consistently retains higher accuracy than the AE baseline at the same number of heads pruned. These results show promise for our greedy approach to head pruning, suggesting that dynamic gradient-based scoring offers an interesting alternative to static importance measures, and providing a foundation for integrating head pruning with broader model compression strategies.


\section*{Code availability}
Code used to produce the numerical experiments in this work is available at the GitHub repository \url{https://github.com/dionysus23334/Greedy-Gnorm}, Release v2026.02.

\section*{Author Contributions}
\textbf{Yuxi Guo}: Conceptualization, Derivations, Numerical experiments, Computer code, Results interpretation, Writing – original draft, Writing – and review \& editing. \textbf{Paul Sheridan}: Supervision, Results interpretation, Writing – and review \& editing. All authors reviewed the results and approved the final version of the manuscript.

\acks{This work was supported by Mitacs through the Mitacs Globalink Research Internship program.}


\newpage

\appendix
\section{Model Parameter Statistics Before and After Pruning}
\label{app:theorem}
This appendix summarizes the parameter composition and post-pruning statistics of representative transformer models evaluated in this study. Tables~\ref{tab:albert-params}–\ref{tab:xlmr-params} report detailed layer-wise parameter counts, storage sizes (in megabytes), and accuracy before and after pruning with Greedy-Gnorm. These results illustrate the magnitude of parameter reduction achieved while maintaining competitive accuracy across different architectures.

\begin{table}[!htbp]
\centering
\caption{ALBERT parameters before/after pruning with Greedy-Gnorm (MB=megabytes).}
\label{tab:albert-params}
\begin{tabular}{lrrrrrr}
\toprule
\multirow{2}{*} & \multicolumn{3}{c}{Before Pruning} & \multicolumn{3}{c}{After Pruning} \\
\cmidrule(lr){2-4} \cmidrule(lr){5-7}
{Layer} & Params & Share (\%) & MB & Params & Share (\%) & MB \\
\midrule
Model      & 11{,}685{,}891 & 100.00 & 44.58 & 11{,}095{,}491 & 100.00 & 42.33 \\
ALBERT     & 11{,}683{,}584 & 99.98  & 44.57 & 11{,}093{,}184 & 99.98  & 42.32 \\
Embeddings & 3{,}906{,}048  & 33.43  & 14.90 & 3{,}906{,}048  & 35.20  & 14.90 \\
Encoder    & 7{,}186{,}944  & 61.50  & 27.42 & 6{,}596{,}544  & 59.45  & 25.16 \\
Pooler     & 590{,}592      & 5.05   & 2.25  & 590{,}592      & 5.32   & 2.25 \\
Classifier & 2{,}307        & 0.02   & 0.01  & 2{,}307        & 0.02   & 0.01 \\
Weight     & 2{,}304        & 0.02   & 0.01  & 2{,}304        & 0.02   & 0.01 \\
Bias       & 3              & 0.00   & 0.00  & 3              & 0.00   & 0.00 \\
\midrule
Accuracy   & \multicolumn{3}{c}{84.48\%} & \multicolumn{3}{c}{77.76\%} \\
\bottomrule
\end{tabular}
\end{table}

\begin{table}[!htbp]
\centering
\caption{RoBERTa parameters before/after pruning with Greedy-Gnorm (MB=megabytes).}
\label{tab:roberta-params}
\begin{tabular}{lrrrrrr}
\toprule
\multirow{2}{*} & \multicolumn{3}{c}{Before Pruning} & \multicolumn{3}{c}{After Pruning} \\
\cmidrule(lr){2-4} \cmidrule(lr){5-7}
{Layer} & Params & Share (\%) & MB & Params & Share (\%) & MB \\
\midrule
Model      & 355{,}361{,}794 & 100.00 & 1355.60 & 291{,}089{,}474 & 100.00 & 1110.42 \\
RoBERTa    & 354{,}310{,}144 & 99.70  & 1351.59 & 290{,}037{,}824 & 99.64  & 1106.41 \\
Embeddings & 52{,}000{,}768  & 14.63  & 198.37  & 52{,}000{,}768  & 17.86  & 198.37 \\
Encoder    & 302{,}309{,}376 & 85.07  & 1153.22 & 238{,}037{,}056 & 81.77  & 908.04 \\
Classifier & 1{,}051{,}650   & 0.30   & 4.01    & 1{,}051{,}650   & 0.36   & 4.01 \\
Dense      & 1{,}049{,}600   & 0.30   & 4.00    & 1{,}049{,}600   & 0.36   & 4.00 \\
Out\_proj  & 2{,}050         & 0.00   & 0.01    & 2{,}050         & 0.00   & 0.01 \\
\midrule

Accuracy   & \multicolumn{3}{c}{87.80\%} & \multicolumn{3}{c}{86.40\%} \\

\bottomrule
\end{tabular}
\end{table}

\begin{table}[!htbp]
\centering
\caption{XLM-RoBERTa parameters before/after pruning with Greedy-Gnorm (MB=megabytes).}
\label{tab:xlmr-params}
\begin{tabular}{lrrrrrr}
\toprule
\multirow{2}{*} & \multicolumn{3}{c}{Before Pruning} & \multicolumn{3}{c}{After Pruning} \\
\cmidrule(lr){2-4} \cmidrule(lr){5-7}
{Layer} & Params & Share (\%) & MB & Params & Share (\%) & MB \\
\midrule
Model      & 278{,}059{,}028 & 100.00 & 1060.71 & 257{,}395{,}028 & 100.00 & 981.88 \\
RoBERTa    & 277{,}453{,}056 & 99.78  & 1058.40 & 256{,}789{,}056 & 99.76  & 979.57 \\
Embeddings & 192{,}398{,}592 & 69.19  & 733.94  & 192{,}398{,}592 & 74.75  & 733.94 \\
Encoder    & 85{,}054{,}464  & 30.59  & 324.46  & 64{,}390{,}464  & 25.02  & 245.63 \\
Classifier & 605{,}972       & 0.22   & 2.31    & 605{,}972       & 0.24   & 2.31 \\
Dense      & 590{,}592       & 0.21   & 2.25    & 590{,}592       & 0.23   & 2.25 \\
Out\_proj  & 15{,}380        & 0.01   & 0.06    & 15{,}380        & 0.01   & 0.06 \\
\midrule
Accuracy   & \multicolumn{3}{c}{99.73\%} & \multicolumn{3}{c}{90.97\%} \\
\bottomrule
\end{tabular}
\end{table}

\section{Collapsed Gradient Handling after Structural Pruning}
\label{app:collapsed-gradient}

Our pruning implementation is based on the \textsc{TextPruner} framework, which performs structural pruning of transformer attention heads rather than simple weight masking. When a head is pruned, \textsc{TextPruner} physically removes its corresponding query, key, value, and output projection blocks from the model. As a result, the associated gradient tensors also shrink in dimensionality so that each pruned head effectively collapses the gradient matrix by eliminating its corresponding sub-blocks.

After pruning, the dimensionality of per-head gradient blocks decreases as certain attention heads are removed. This collapse poses a challenge for maintaining consistent gradient statistics across pruning iterations. 
The reduced vectors cannot be directly compared to their pre-pruning counterparts. Their length and positional correspondence within the layer both change after pruning. To ensure that gradient-based importance measures remain meaningful over time, we expand each shrunken gradient vector back to its original dimensionality using the binary mask $M$, inserting zeros at pruned positions. This reconstruction preserves the spatial alignment of gradient entries across pruning steps, allowing the expected gradient-norm signal $S(n)$ to remain comparable and stable as the model structure evolves.

\section{Attention Entropy Rectification}
\label{app:epsilon-discussion}
AE is susceptible to numerical underflow because attention probabilities often become extremely small when the sequence length is large. This section formalizes our $\varepsilon$-rectified entropy variants and explains why the chosen rectification preserves the relative head ordering used for pruning.

\subsection{Epsilon Fine-Tuning}
When the number of tokens is large, attention values are dispersed across many positions, and $log(0)$ may appear in the entropy computation. 
To address this issue, we introduce $\varepsilon$-rectified variants of the AE:
\begin{align}
A(a_1,a_2,\ldots,a_n) &= -\sum_{i=1}^{n} a_i \log a_i, \label{eq:AE-A}\\[4pt]
B(a_1,a_2,\ldots,a_n) &= -\sum_{i=1}^{n} a_i \log(a_i+\varepsilon), \label{eq:AE-B}\\[4pt]
C(a_1,a_2,\ldots,a_n) &= -\sum_{i=1}^{n} (a_i+\varepsilon)\log(a_i+\varepsilon). \label{eq:AE-C}
\end{align}

By construction, $A$ is the original entropy, while $B$ and $C$ are its $\varepsilon$-rectified forms. 
Their relationships can be expressed as:
\begin{equation}
C - A = (C - B) + (B - A).
\label{eq:AE-rel}
\end{equation}
Here $\varepsilon$ is a small positive constant such that $a_i+\varepsilon<1$. 
It follows that $C - B > 0$ and $B - A < 0$. 
The term $B - A < 0$ is straightforward, so we omit its proof.  
Given that $\sum_i a_i = 1$ and $\varepsilon \ll \tfrac{1}{2} \leq \tfrac{n-1}{n}$, 
we can show:
\begin{equation}
\left(\prod_i (a_i+\varepsilon)\right)^{\frac{1}{n}} < \frac{n\varepsilon + 1}{n} < 1,
\label{eq:AE-ineq}
\end{equation}
which implies $C - B > 0$.  

In practice, we adopt $C$ as our entropy variant because it is always greater than $B$ and thus provides stronger numerical stability (no NaNs) while maintaining head ranking consistency.  
When $A$ suffers from underflow, $B$ eliminates the log-zero issue, but its entropy value itself may still approach zero.  
Using $C$ ensures that the entropy computation remains numerically stable even when $a_i$ is extremely small.

\subsection{Invariant Pruning Order}
\label{app:invariant-order}
Let 
\begin{equation}
F(a_1,a_2,\ldots,a_n)=\sum_{i=1}^{n}a_i\log a_i
\label{eq:F-def}
\end{equation}
be continuous on $\mathbb{R}^+$.  
Suppose two attention distributions $P_1=(x_1,\ldots,x_n)$ and $P_2=(y_1,\ldots,y_n)$ satisfy $\sum_i x_i = \sum_i y_i = 1$, and define
\begin{equation}
\delta = F(P_2) - F(P_1) > 0.
\label{eq:delta-def}
\end{equation}
We now examine whether this inequality still holds after $\varepsilon$-rectification.  
Define 
\begin{align}
P_1'&=(x_1+\varepsilon,\ldots,x_n+\varepsilon), \label{eq:P1-prime}\\
P_2'&=(y_1+\varepsilon,\ldots,y_n+\varepsilon), \label{eq:P2-prime}
\end{align}
and let 
\begin{equation}
\delta' = F(P_2') - F(P_1').
\label{eq:delta-prime}
\end{equation}
We ask: does $\delta' > 0$ still hold?

Consider the segment connecting $P_1$ and $P_2$ in $n$-dimensional space, parameterized as
\begin{equation}
\vec{l} = (l_1,l_2,\ldots,l_n)
    = \frac{\vec{P_1P_2}}{\|\vec{P_1P_2}\|}
    = \frac{\vec{P'_1P'_2}}{\|\vec{P'_1P'_2}\|}.
\label{eq:l-def}
\end{equation}
The difference $\delta$ can be expressed as the line integral of the directional derivative of $F$:
\begin{equation}
\delta = F(P_2) - F(P_1)
       = \int_{P_1}^{P_2} \nabla F \cdot \vec{l}\, dl > 0,
\label{eq:delta-integral}
\end{equation}
where the gradient is
\begin{equation}
\nabla F = (\log a_1 + 1, \log a_2 + 1, \ldots, \log a_n + 1).
\label{eq:gradF}
\end{equation}
Thus,
\begin{equation}
\nabla F \cdot \vec{l}
  = \sum_{i=1}^{n} l_i \log a_i.
\label{eq:grad-dot}
\end{equation}

Since $F$ is continuous and smooth on $\mathbb{R}^+$, for sufficiently small $\varepsilon$ the perturbed gradient remains in a local neighborhood of the original:
\begin{equation}
\nabla F' \cdot \vec{l}
   = \sum_{i=1}^{n} l_i \log(a_i+\varepsilon)
   \in N_r(\nabla F \cdot \vec{l}),
\label{eq:grad-perturbed}
\end{equation}
where $r$ controls the allowed deviation in the integral value.  
Consequently,
\begin{equation}
\delta' = F(P_2') - F(P_1') 
        = \int_{P_1'}^{P_2'} \nabla F' \cdot \vec{l}\, dl > 0.
\label{eq:delta-prime-integral}
\end{equation}

This result shows that within a sufficiently small $\varepsilon$-neighborhood, the sign of $\delta$ remains unchanged, and hence the relative ordering of AE values is preserved.  
Therefore, using the rectified form $C$ allows AE-based pruning to remain stable and monotonic—avoiding numerical underflow without altering the ranking of heads used for pruning.

\vskip 0.2in
\bibliography{bibliography}

\end{document}